\documentclass[letterpaper, 10 pt, conference]{ieeeconf}  % Comment this line out if you need a4paper
\IEEEoverridecommandlockouts                              % This command is only needed if 
\usepackage{amsmath} % assumes amsmath package installed
\usepackage{algorithm}
\usepackage{algorithmic}
\usepackage{bbm}
\usepackage{dsfont}
\usepackage{xcolor}
\usepackage{amssymb}
\usepackage{graphicx}
\usepackage{mathtools}
\usepackage{amsmath}
\usepackage{gensymb}
\usepackage{float}
\usepackage{multirow}
\usepackage{makecell}
\usepackage{mathtools}
\usepackage{hyperref}
\hypersetup{
    colorlinks=true,
    linkcolor=blue,
    filecolor=magenta,      
    urlcolor=blue,
    pdftitle={Overleaf Example},
    pdfpagemode=FullScreen,
    }

\usepackage{tabularx}
    \newcolumntype{L}{>{\raggedright\arraybackslash}X}
\usepackage[utf8]{inputenc}
\usepackage[english]{babel}
\usepackage{amsthm}

\usepackage{subcaption}

\newcommand{\no}{\noindent}

\title{\LARGE \bf CROSS-GAiT: Cross-Attention-Based Multimodal Representation Fusion for Parametric Gait Adaptation in Complex Terrains
}

\newcommand{\rev}[1]{\textcolor{black}{#1}}

\author{
    Gershom Seneviratne$^{1}$, 
    Kasun Weerakoon$^{1}$, 
    Mohamed Elnoor$^{1}$,  
    Vignesh Rajgopal$^{3}$, \\
    Harshavarthan Varatharajan$^{3}$,  
    Mohamed Khalid M Jaffar$^{4}$, 
    Jason Pusey$^{5}$, 
    and Dinesh Manocha$^{1,2}$
    \thanks{$^{1}$ Authors are with Dept. of Electrical and Computer Engineering, University of Maryland, College Park, MD, USA. {\tt\footnotesize
    gershom@umd.edu
    kasunw@umd.edu,  melnoor@umd.edu,
    dmanocha@umd.edu}
2    }
    \thanks{$^{2}$ Author is with Dept. of Computer Science, University of Maryland, College Park, MD, USA. {\tt\footnotesize dmanocha@umd.edu}
    }
    \thanks{$^{3}$ Author is with A. James Clark School of Engineering, University of Maryland, College Park, MD, USA. {\tt\footnotesize vigneshr@umd.edu, 
    harshav@umd.edu}
    }
    \thanks{$^{4}$ Author is with Dept. of Aerospace Engineering, University of Maryland, College Park, MD, USA. {\tt\footnotesize khalid26@umd.edu}
    }
    \thanks{$^{5}$ Author is with U.S. Army Research Laboratory (ARL). {\tt\footnotesize jason.l.pusey.civ@army.mil}
    }
    \thanks{This work was supported in part by ARO Grants  W911NF2310046, W911NF2310352, and U.S. Army Cooperative Agreement W911NF2120076.
    }
}

\newcommand{\cmnt}[1]{\textcolor{black}{#1}}

\newcommand{\ours}{CROSS-GAiT} % alias for the method name

\begin{document}
\raggedbottom
\maketitle
\thispagestyle{empty}
\pagestyle{empty}

%%%%%%%%%%%%%%%%%%%%%%%%%%%%%%%%%%%%%%%%%%%%%%%%%%%%%%%%%%%%%%%%%%%%%%%%%%%%%%%%
% \vspace{-40pt}
\begin{abstract}

We present \ours{}, a novel algorithm for quadruped robots that uses Cross Attention to fuse terrain representations derived from visual and time-series inputs; including linear accelerations, angular velocities, and joint efforts. These fused representations are used to continuously adjust two critical gait parameters (step height and hip splay), enabling adaptive gaits that respond dynamically to varying terrain conditions. 
To generate terrain representations, we process visual inputs through a masked Vision Transformer (ViT) encoder and time-series data through a dilated causal convolutional encoder. The Cross Attention mechanism then selects and integrates the most relevant features from each modality, combining terrain characteristics with robot dynamics for informed gait adaptation. This fused representation allows \ours{} to continuously adjust gait parameters in response to unpredictable terrain conditions in real-time. We train \ours{} on a diverse set of terrains including asphalt, concrete, brick pavements, grass, dense vegetation, pebbles, gravel, and sand and validate its generalization ability on unseen environments. Our hardware implementation on the Ghost Robotics Vision 60 demonstrates superior performance in challenging terrains, such as high-density vegetation, unstable surfaces, sandbanks, and deformable substrates.
We observe at least a 7.04\% reduction in IMU energy density and a 27.3\% reduction in total joint effort, which directly correlates with increased stability and reduced energy usage when compared to state-of-the-art methods. Furthermore, \ours{} demonstrates at least a 64.5\% increase in success rate and a 4.91\% reduction in time to reach the goal in four complex scenarios. Additionally, the learned representations perform 4.48\% better than the state-of-the-art on a terrain classification task.

\end{abstract}

\section{Introduction} \label{sec:intro}

Recent advancements in quadruped robot platforms have been used for autonomous navigation in challenging outdoor environments, such as rocky surfaces, deformable terrains, and densely vegetated areas where wheeled or tracked robots do not work well~\cite{miki2022learning, frey2022locomotion, frey2023fast, sathyamoorthy2024mim, agarwal2023legged, karnan2023sterling}. To achieve consistent performance and to navigate efficiently, quadruped robots must rely on accurate terrain understanding, which is essential for maintaining stability, reducing energy consumption, and supporting long missions in complex environments \cite{fu2021minimizing, roy2012effects, hutter2016anymal}.

\begin{figure}[!t]
      
      % \linespread{1.0}
      \centering
      \includegraphics[width=\columnwidth]{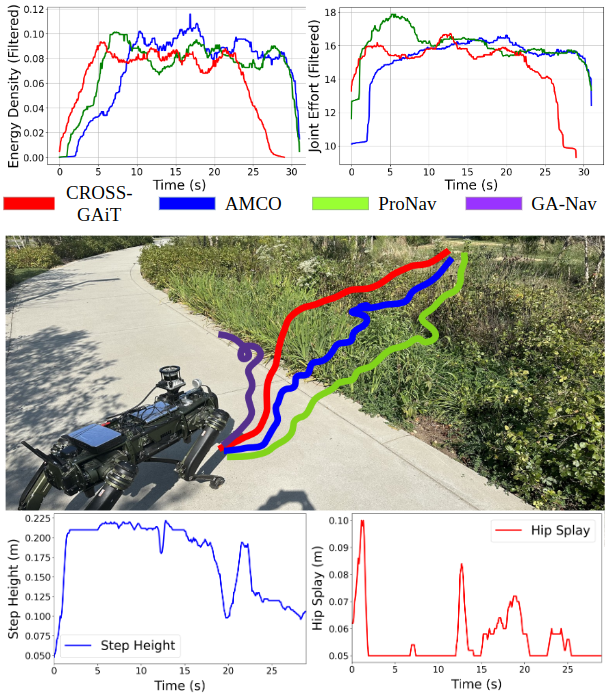}
      \caption {\small{Comparison of \ours{} with state-of-the-art methods while navigating a Ghost Robotics Vision 60 robot through a complex scenario that contains a paved concrete path, thick bushes, grass, and pebbles: \rev {\ours{} continuously adapts gait parameters (step height and hip splay) based on multi-sensory observations and outperforms all other comparison methods \cite{elnoor2024amco,elnoor2024pronav,guan2022ga} in terms of success rate and time to reach the goal.} \textbf{[TOP]} two graphs represent IMU joint effort and IMU energy density (both median-filtered to reduce noise artifacts) over time, illustrating the reduced energy consumption and improved stability achieved by \ours{} in comparison to the other methods. \textbf{[BOTTOM]} two graphs present the continuous variation of gait parameters according to different terrain conditions. For instance, \ours{} uses high steps in dense vegetation to reduce leg entrapment while increasing hip splay during instabilities or sinkage, even when the camera is occluded by tall grass.} }
      \label{fig:cover-image}
      \vspace{-15pt}
\end{figure}
Exteroceptive sensors, such as cameras and LiDARs, play a key role in providing the terrain understanding needed for navigating such unstructured environments. Numerous approaches leverage these sensors \cite{yu2021visual, guan2022ga, li2023seeing, sathyamoorthy2024mim}, combined with learning-based methods and multi-modal fusion techniques, to perform terrain segmentation, classification, and generate occupancy maps for reliable traversability estimates \cite{xue2018deep, guan2022ga, sathyamoorthy2023vern, weerakoon2022graspe}. However, current methods mostly rely on manually annotated terrain datasets, which are labor-intensive, time-consuming, and prone to human bias.
%limiting their real-world applicability.

%Proprioception
Proprioceptive sensors assess instabilities on various terrains, enhancing terrain understanding \cite{dey2022prepare, elnoor2024pronav, kumar2021rma} and allowing robots to identify unstructured regions within visually uniform areas. However, relying solely on proprioception requires the robot to traverse the terrain, limiting proactive avoidance of mostly unstable areas \cite{dey2022prepare, elnoor2024pronav, kumar2021rma}. To address this, prior methods have integrated proprioceptive and exteroceptive data using heuristics~\cite{elnoor2024amco}. However, these methods may not be able to capture complex relationships between multi-modal datasets due to their predefined and fixed nature. Furthermore, 
self-supervised and unsupervised learning approaches have been used to fuse latent representations from proprioceptive and exteroceptive inputs and generate trajectories that can navigate through traversable regions~\cite{sikand2022visual, castro2023does, wellhausen2019should, kahn2021badgr, yao2022rca, liu2024contrastive}.

Effective gait adaptation is critical in complex scenarios where avoiding unstable regions is infeasible, such as when a quadruped robot encounters dense vegetation, sandy or rocky terrains. Using a fixed gait in such conditions may lead to increased energy consumption and instability \cite{matthis2018gaze, lewis-gait, fu2021minimizing, roy2012effects}. Several methods have been proposed to switch between pre-defined gaits based on estimated traversability calculated using simple rules~\cite{elnoor2024amco, elnoor2024pronav}, but their application has been limited.
%While these approaches offer a degree of adaptability, they often lead to less effective gait adjustments due to the limited gait options available. 

%and complicating the guarantee of stability in real-world applications.

\textbf{Main contributions:} To address these limitations, we present \textbf{\ours{}} (\textbf{Cross}-Attention-Based Multi-modal Representation Fusion for parametric \textbf{G}ait \textbf{A}daptation \textbf{i}n Complex \textbf{T}errains), a novel approach that fuses terrain representations from visual, IMU, and proprioceptive inputs using a cross-attention transformer network. Our fused representation is leveraged to \rev{continuously} adjust gait parameters that adapt to the terrain, reduce energy consumption and increase perceived stability during navigation. The novel components of our work include:

%In addition to these learning techniques, cross-attention offers a powerful mechanism for fusing these two modalities, allowing them to attend to each other's features and produce richer, more comprehensive latent representations \cite{li2024crossfuse, rajan2022cross}. Despite its potential, cross-attention remains relatively underexplored in robotics, particularly for tasks like terrain-aware navigation.

\begin{itemize}

\item \textbf{A novel multi-modal fusion algorithm} that fuses latent representations of visual data with time-series data (linear accelerations, angular velocities, and proprioception) using cross-attention, resulting in a comprehensive latent representation of the environment. Trained with supervised contrastive loss, \ours{} achieves 98.45\% terrain classification accuracy when evaluated on test data from the dataset described in Section \hyperref[sec:Dataset]{V-B}, outperforming state-of-the-art and classical MLP-based fusion methods at least by 3.18\% in terms of terrain classification accuracy using the learned representations. 

% \item \textbf{A parametric gait adaptation algorithm: } Continuously adjusts two gait parameters (step height and hip splay) to dynamically change the quadruped robot's gait instead of switching between a pre-defined set of available gaits. This ctinouos and dynamic gait parameter adapatation lead to 7.04\% of improved stability and reduced joint effort by 27.3\%, compared to existing gait adaptation methods.  

\rev{ \item \textbf{A novel parametric gait adaptation algorithm} that continuously adjusts two critical gait parameters (step height and hip splay) enabling smooth and dynamic transitions in the quadruped robot’s gaits. Unlike traditional methods that switch between a predefined set of discrete gaits, our approach allows for real-time and continuous gait modulation based on environmental conditions and proprioceptive feedback. This continuous adaptation improves stability by 7.04\% and reduces joint effort by 27.3\% compared to existing gait adaptation techniques.}

% \item  \textbf{A real-time implementation: } Our model system is lightweight by design to run on edge hardware. Hence in our experiment setup, \ours{} runs around 60 Hz on a mobile computer an Intel NUC 11 with an Intel i7 CPU and NVIDIA RTX 2060 GPU. We demonstrate \ours{} on a Ghost Robotics Vision60, evaluating its performance in complex outdoor scenarios on top of an exteroceptive navigation algorithm \cite{sathyamoorthy2024mim}. 

\rev{ \item  \textbf{A real-time implementation: } Our model is designed to be lightweight for real-time execution on edge hardware. \ours{} achieves a processing rate of approximately 60 Hz on a mobile computing platform equipped with an Intel NUC 11 (Intel i7 CPU, NVIDIA RTX 2060 GPU). We validate \ours{} on the Ghost Robotics Vision60 quadruped, demonstrating its effectiveness in complex outdoor environments using an exteroceptive navigation algorithm \cite{sathyamoorthy2024mim}, showcasing robust performance in real-world deployment scenarios.}

\end{itemize}
\section{Related Work} \label{sec:related_work}

In this section, we give a brief overview of existing work in perception methods for terrain understanding, gait change mechanisms for navigation in complex environments, and cross attention based multi-modal fusion. 

\subsection{Perception Methods for Terrain Understanding}

The notion of incorporating vision and LiDAR-based techniques has been widely used for terrain understanding in outdoor settings. Semantic segmentation and image classification help identify navigable terrains, providing inputs for planning and control algorithms \cite{xue2018deep, schilling2017geometric, sathyamoorthy2023vern}. For example, Guan et al. \cite{guan2022ga} use pixel-level segmentation to update traversability costs, while Fahmi et al. \cite{fahmi2022vital} employ vision-based methods for pose adaptation and foothold selection. However, vision-based methods struggle in varying lighting and occluded environments like dense vegetation \cite{aladem2019evaluation}. LiDAR-based approaches \cite{sathyamoorthy2024mim} and vision-LiDAR combinations \cite{weerakoon2022graspe} improve terrain understanding but still face challenges in detecting deformability and occluded obstacles, which may lead to instability.

Proprioceptive sensory data, such as joint forces and inertial measurements, are widely used for instability detection during navigation \cite{dey2022prepare, elnoor2024pronav}. Dey et al. \cite{dey2022prepare} predict slippage using time-series proprioception data, while Kumar et al. \cite{kumar2021rma} propose a standalone proprioceptive reinforcement learning-based navigation algorithm for outdoor terrains. However, proprioceptive methods struggle to anticipate future terrains, limiting their utility \cite{matthis2018gaze}.
To overcome this, studies have fused proprioceptive data with LiDAR or visual inputs~\cite{elnoor2024amco}.
%explored heuristic rule-based methods combining segmentation and proprioceptive inputs, but these often fail to capture complex multimodal relationships.

\begin{figure*}[t] % The figure* environment makes the figure span both columns
\centering
\includegraphics[width=\textwidth]{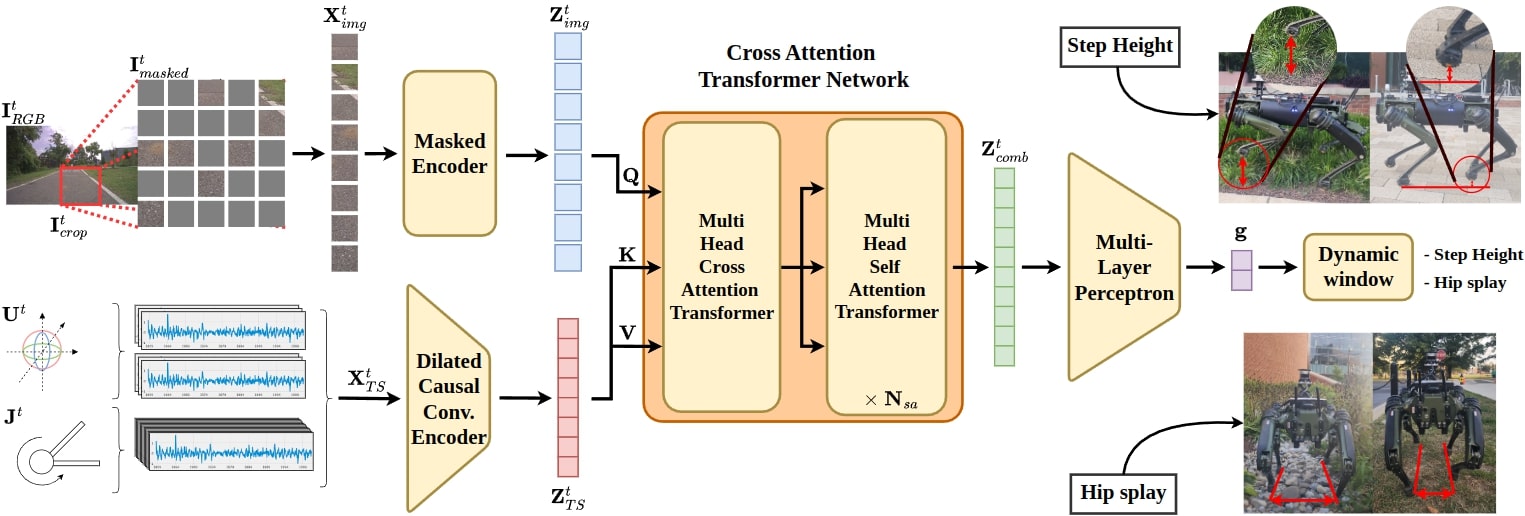} % Replace with your image file
\caption{\small{Overall architecture of \ours{}: Our method utilizes a vision transformer-based masked autoencoder for image data and a dilated causal convolutional encoder for IMU and joint effort data to generate latent representations. These are combined using a cross-attention transformer network. The combined latent representation is then passed through a multilayer perceptron-based regressor to obtain the final output: a set of gait parameters, including step height and hip splay. A dynamic window mechanism is applied to ensure smooth gait transitions during terrain navigation. The overall system runs around 60 Hz on the implemented system. }}
\label{fig:arch} % Label for referencing the image
\vspace{-15pt}
\end{figure*}

Other methods focus on self-supervised traversability cost learning using latent representations of proprioceptive and exteroceptive inputs \cite{sathyamoorthy2022terrapn, sikand2022visual, castro2023does, wellhausen2019should, kahn2021badgr, yao2022rca}. Karnan et al.~\cite{karnan2023sterling} use VICReg loss \cite{bardes2021vicreg} to learn robust terrain representations, enabling successful outdoor navigation based on user terrain preference.  While these methods use terrain representations primarily for traversability cost estimation, our work leverages these representations for adaptive gait changes, enabling navigation through complex terrains.

\subsection{Gait Change Mechanisms for Navigation in Complex Environments}

Gait adaptation helps maintain quadraped stability and reduce energy consumption in complex environments \cite{matthis2018gaze, lewis-gait}. Several studies have examined switching between predefined gaits for stability and energy efficiency~\cite{elnoor2024amco, elnoor2024pronav}.

Reinforcement learning techniques have been employed to train locomotion policies that estimate potential foothold positions, enabling robots to navigate complex terrains with greater efficiency \cite{zhu2024cross, agarwal2023legged, loquercio2023learning, yu2021visual, miki2022learning, fu2022coupling}.  These methods can result in significant improvements in the efficiency and adaptability of robot locomotion. In particular, Miki et al.~\cite{miki2022learning} successfully autonomously navigated a robot on a hiking trail using a reinforcement learning-based policy that handled complex scenarios. 
%Although these methods can demonstrate significant improvements in locomotion, 
Most of these methods rely on high-fidelity simulations, making sim2real transfer difficult. As a result, it is challenging to ensure stability and robustness in real-world deployments on complex terrains.

\subsection{Cross-Attention Multi-modal Fusion}

Cross-attention mechanisms have demonstrated significant success in fusing information from multiple modalities across various domains for useful downstream tasks such as object detection~\cite{bahaduri2023multimodal}, image and sentence matching~\cite{wei2020multi}, information fusion~\cite{li2024crossfuse}, emotion detection~\cite{praveen2024recursive}, recommendor systems~\cite{limultimodalcatt} and soft robot manipulation~\cite{zhao2024bronchocopilot}. For example, BronchoCopilot, a multi-modal reinforcement learning agent designed for autonomous robotic bronchoscopy, employs cross-attention mechanisms to fuse visual information from the bronchoscope camera with the robot's estimated pose. This fusion enables the system to navigate complex airway environments more effectively.

In contrast to these existing methods, we fuse proprioceptive time-series data with exteroceptive visual data to generate a comprehensive terrain representation, enabling dynamic gait parameter adjustments for quadruped robots navigating diverse terrains.

\section{Background}
% \label{sec:background}

\subsection{Notations}

We use the following notations: \(N\) for architectural counts (e.g., layers or blocks), \(n\) and \(m\) for input and batch sizes, respectively, \(J\) for joint effort data, \(I\) for image inputs, \(U\) for IMU data, and \(Z\) for latent representations. \(\mathcal{L}\) represents losses, \(W\) denotes weight matrices, and \(Q\), \(K\), \(V\) are query, key, value vectors in attention mechanisms, respectively. \(d\) refers to dimensions, \(t\) to time, \(g\) to gait parameters, and \(\theta\) to terrains.

\subsection{Representation Learning}

Representation learning automatically discovers useful features from raw data, allowing machines to perform tasks without manual feature extraction \cite{bengio2013representation}. For terrain-aware navigation, effective representations enhance understanding of the terrain and robot dynamics, improving decision-making for gait adaptation and trajectory cost estimation. Modern approaches use both visual and time-series inputs to capture the spatial and temporal dynamics of the environment and the robot's interaction with it.

\subsection{Masked Autoencoders for Visual Representation Learning}

\label{sec:MAE}

Masked autoencoders are self-supervised models that learn robust terrain representations by reconstructing occluded portions of an input image.  The encoder generates latent representations, while the decoder reconstructs the original image. By training the model to predict masked regions, the encoder extracts high-level features that capture key patterns and structures, representing terrain features in a lower-dimensional space. 

We implement the masked autoencoder (MAE) proposed by He et al. \cite{he2022masked}, using a ViT backbone \cite{dosovitskiy2020image} for image reconstruction, and extract the encoder to generate terrain representations for downstream tasks.

\subsection{Dilated Causal Convolutional Encoder for Time-Series Representation Learning}

\label{sec:DConvol}

Dilated causal convolutional encoders are neural networks for sequential data, using dilated convolutions to capture patterns over varying time scales \cite{ma2021ecg}. Unlike standard convolutions, they expand the receptive field by introducing gaps between filter elements, allowing deeper networks to capture long-range dependencies without increasing model complexity. Causal convolutions ensure that outputs are influenced only by current or past inputs, preserving the sequential nature of time-series data.

We adopt the dilated causal convolutional encoder by Franceschi et al. \cite{franceschi2019unsupervised} to extract temporal terrain representations from IMU readings and joint effort measurements for downstream tasks.

\subsection{Cross-Attention Transformer Mechanism}

\label{sec:CAtt}

\rev{Cross-attention fuses information from multiple modalities by allowing one modality to attend to relevant features from another. This allows a model to selectively emphasize relevant features from each modality, reducing redundancy and enhancing contextual understanding. In multimodal learning, it enables the model to focus on the most informative aspects of each modality, such as visual and time-series inputs \cite{li2024crossfuse, rajan2022cross}. Queries, Q from one modality attend to keys K and values V from another, integrating complementary information for richer representations. Other components of the transformer, like add \& norm layers, residual connections, and feedforward networks, remain unchanged \cite{vaswani2017attention}. Overall, unlike concatenation followed by MLPs \cite{karnan2023sterling}, which apply static transformations, cross-attention enables adaptive feature alignment, leading to more robust and informative representations.}
\section{Our Approach: \ours{}} 

We present \ours{}, a method for navigating complex outdoor environments inspired by how humans adapt their gait using visual and proprioceptive feedback. Similar to humans adjusting their gait for stability on diverse terrains, \ours{} ensures stable and energy-efficient gaits for surfaces like grass, sand, asphalt, and rocky terrains. Our approach focuses on four key aspects: generating latent representations from visual data, generating representations from time-series data, fusing these multi-modal inputs via cross-attention, and using the fused representation to generate adaptive gait parameters. The overall architecture is outlined in \cmnt{Figure \ref{fig:arch}}.

\subsection{Latent Representations of Visual Data}

Given an RGB image at time \( t \) (\(\mathbf{I}^t_{\text{RGB}} \in \mathbb{Z}^{H \times W \times 3}\)), where \( H \) is the height, \( W \) is the width, and 3 represents the color channels, we crop a square-shaped sub-image \(\mathbf{I}^t_{\text{crop}} \in \mathbb{Z}^{n_i \times n_i \times 3}\) from the bottom center of \(\mathbf{I}^t_{\text{RGB}}\), as shown in \cmnt{Figure \ref{fig:arch}}. This sub-image corresponds to the immediate terrain the robot is about to traverse, and \( n_i \) is the input size required by the MAE model. Before passing \(\mathbf{I}^t_{\text{crop}}\) through the encoder of the MAE, it is masked by dividing it into 16x16 patches and randomly masking a percentage of these patches. The remaining visible patches \(\mathbf{X}^t_{\text{img}} \in \mathbb{R}^{n_v \times (16 \times 16 \times 3)}\) are then passed to the ViT backbone, which generates a latent representation of the robot's visual inputs, \(\mathbf{Z}^t_{\text{img}} \in \mathbb{R}^{n_v \times d_e}\), where \( n_v \) is the number of unmasked patches and \( d_e \) is the embedding dimension. This representation captures the terrain features in a compact form for use in downstream tasks.

We initialize the model with pre-trained weights from \cite{he2022masked} and fine-tune it on our custom dataset (described in Section \hyperref[sec:Dataset]{V-B}), ensuring that the learned representations are well-suited to the terrains the robot would encounter.
 
\vspace{-3pt}
\begin{equation}
\label{eq:mae}
\mathcal{L}_{\text{MAE}} = \frac{1}{|\mathcal{M}|} \sum_{i \in \mathcal{M}} \left\| \mathbf{I}_i - \hat{\mathbf{I}}_i \right\|_2^2
\end{equation}

The Mean Squared Error (MSE) loss (\ref{eq:mae}) between the original image patches \( \mathbf{I}_i \) and the reconstructed patches \( \hat{\mathbf{I}}_i \) in the masked regions \( \mathcal{M} \) is used to fine-tune the model\cite{he2022masked}.

\subsection{Latent Representation of Time Series Data}
We combine IMU data and joint effort data to form the input time-series data \(\mathbf{X}^t_{\text{TS}} \in \mathbb{R}^{n_{\text{ch}} \times d_{\text{TS}}}\), where \(n_{\text{ch}}\) is the number of channels from both sources, and \(d_{\text{TS}}\) represents the temporal dimension of \(\mathbf{X}^t_{\text{TS}}\). We generate latent representations (\(\mathbf{Z}^t_{\text{TS}} \in \mathbb{R}^{1 \times d_m}\)) for \(\mathbf{X}^t_{\text{TS}}\) using the dilated causal convolutional model, discussed in Section~\hyperref[sec:DConvol]{III-D}. Here \(d_m\) is the dimension of the time series latent representation. 

The model is trained on the dataset described in Section \hyperref[sec:Dataset]{V-B}, ensuring it effectively learns terrain features from the time-series data.

\vspace{-14pt}
\begin{equation}
\label{eq:triplet}
\mathcal{L}_{\text{triplet}} = -\log \sigma\left( y_{\text{anc}}^\top \cdot y_{\text{pos}} \right) - \sum_{k=1}^{K} \log \sigma\left( -y_{\text{anc}}^\top \cdot y_{\text{neg}, k} \right)
\end{equation}

The triplet loss (\ref{eq:triplet}) optimizes model training by differentiating between positive samples (\(y_{\text{pos}}\)) and negative samples (\(y_{\text{neg}}\)) based on their similarity to anchor samples (\(y_{\text{anc}}\)) \cite{schroff2015facenet}. This loss function enables the model to learn and encode differences in time-series data from various terrains more effectively.

\begin{figure}[!t]
      \centering
      \includegraphics[width=\columnwidth]{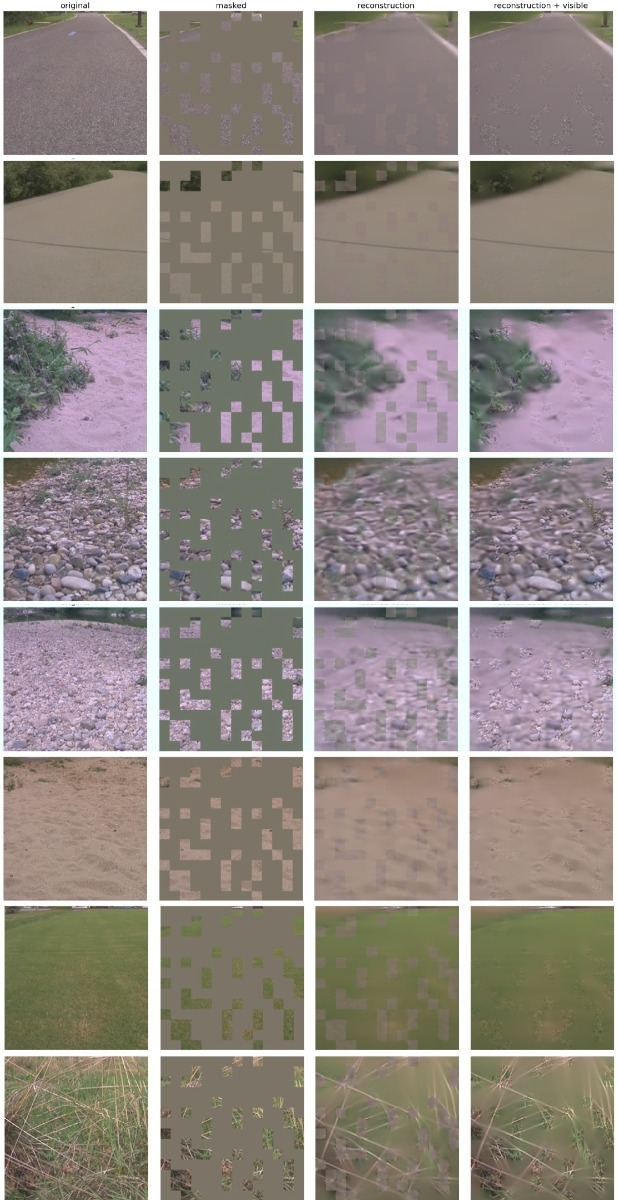}
      \caption {\small{Image reconstruction outputs from the masked autoencoder (MAE) trained for feature encoding. 75\% of the input images are masked and fed into the MAE for reconstruction. The fine-tuned MAE captures fine-grained details such as dry leaves and the granularity of sand/pebbles. During inference, only the encoder is used to obtain latent representations that encode image features for complex terrains.}}
      % \red{get a better photo}}
      \label{fig:MAE_samples}
      \vspace{-15pt}
\end{figure}

\subsection{Combining Latent Representations of Visual Data and Time Series Data}

We use a multi-head cross-attention transformer, as described in Section \hyperref[sec:CAtt]{III-E}, to extract interactions between the multi-modal latent representations \( \mathbf{Z}^t_{\text{TS}} \) and \( \mathbf{Z}^t_{\text{img}} \). We generate the query vector \( \mathbf{Q} \in \mathbb{R}^{1 \times n_h \cdot d_k} \) from the time series latent representation \( \mathbf{Z}^t_{\text{TS}} \) by multiplying it with a weight matrix \( \mathbf{W}_Q \in \mathbb{R}^{d_m \times n_h \cdot d_k} \), where \( d_k \) is the dimension of each attention head and \( n_h \) is the number of heads. We obtain the key matrix \( \mathbf{K} \in \mathbb{R}^{(n_v + 1) \times n_h \cdot d_k} \) and the value matrix \( \mathbf{V} \in \mathbb{R}^{(n_v + 1) \times n_h \cdot d_v} \) by multiplying \( \mathbf{Z}^t_{\text{img}} \) with the corresponding weight matrices \( \mathbf{W}_K \in \mathbb{R}^{d_e \times n_h \cdot d_k} \) and \( \mathbf{W}_V \in \mathbb{R}^{d_e \times n_h \cdot d_v} \). Here, \( d_v \) is the dimension of the value vector.
\vspace{-14pt}

\begin{equation}
\label{eq:CATT}
\text{Cross-Attention}(Q, K, V) = \text{softmax} \left( \frac{QK^\top}{\sqrt{d_k}} \right) V
\end{equation}

We pass the result of the cross-attention operation (\ref{eq:CATT}), \( \mathbf{Z}^t_{\text{imd}} \in \mathbb{R}^{1 \times d_v} \), through the rest of the cross-attention block, including the add \& norm layer, and a feedforward network, which introduces nonlinearity and refines the fused representation. This output is fed into \( N_{sa} \) self-attention transformer blocks to produce the final output \( \mathbf{Z}^t_{\text{comb}} \in \mathbb{R}^{1 \times d_v} \). The cross-attention fuses relevant features from both modalities, while the self-attention blocks capture deeper dependencies and interactions between the image and time-series data.

\begin{equation}
\label{eq:scl}
\mathcal{L}_{\text{SCL}} = \sum_{i=1}^{N} \frac{-1}{|P(i)|} \sum_{p \in P(i)} \log \frac{\exp(\mathbf{z}_i \cdot \mathbf{z}_p / \tau)}{\sum_{a \in A(i)} \exp(\mathbf{z}_i \cdot \mathbf{z}_a / \tau)}
\end{equation}

We use the dataset described in Section \hyperref[sec:Dataset]{V-B} to train the Cross-Attention transformer network with supervised contrastive loss (\ref{eq:scl}).  This loss pulls together representations of positive samples \( p \in P(i) \) (from the same class as anchor sample \( i \)) while pushing apart representations of other samples \( a \in A(i) \) (both positive and negative examples) \cite{khosla2020supervised}. The temperature parameter \( \tau \) controls the separation of negative classes. The loss is averaged over all \( N \) samples, encouraging the model to learn discriminative features for different terrain types. During the training of the Cross-Attention transformer network, the upstream networks are frozen to retain the representations learned during their initial training.
\vspace{-3pt}

\begin{equation}
\label{eq:cost}
\mathcal{C}(\mathbf{g}, \theta) = e^{\alpha \cdot \tilde{\omega}_t(\mathbf{g}, \theta) + \beta \cdot \tilde{a}_z(\mathbf{g}, \theta) + \gamma \cdot \tilde{J}_{\text{effort}}(\mathbf{g}, \theta)}
\end{equation}

We define a gait parameter cost function (\ref{eq:cost}) to identify parameter combinations that minimize energy consumption and increase stability for the terrains in the training dataset (Section \hyperref[sec:Dataset]{V-B}). The parameter pair with the lowest cost for each terrain is selected to generate the gait parameter labels used for training. In this cost function, \(\tilde{\omega}_t\), \(\tilde{a}_z\), \(\tilde{J}_{\text{effort}}\), \(\alpha\), \(\beta\), and \(\gamma\) represent the normalized angular velocity, linear acceleration along the z axis, RMS joint effort, and their corresponding weights. Parameter selection for training is detailed in Section \hyperref[sec:Dataset]{V-B}.

\begin{figure}[t]
      \centering
      \hspace*{-0.25cm}\includegraphics[width=1.04\columnwidth]{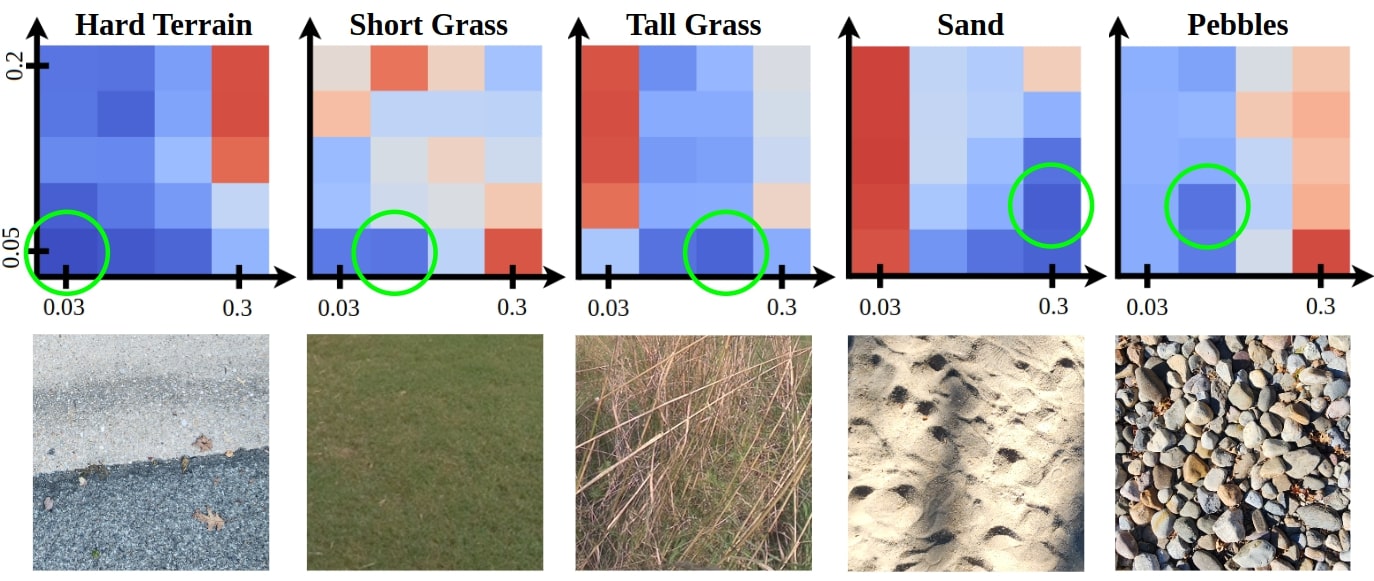}
      \caption {\small{\textbf{[TOP]:} Illustration of the cost values evaluated for different gait parameter pairs using the cost function defined in Equation \ref{eq:cost}. The \textbf{\textit{x axis}} denotes step height and the \textbf{\textit{y axis}} represents hip splay. The circled parameter pairs represent the selected labels for training \textbf{\textit{the regression}}.  \textbf{[BOTTOM]:} Sample terrain images corresponding to each surface type in the above plots. Paramater pairs that lead to the lowest cost (dark blue) in each terrain are highlighted in green circles. }}
      \label{fig:Params}
      \vspace{-20pt}
\end{figure}

\subsection{Model for Gait Parameter Generation}

We pass the combined latent representation \( \mathbf{Z}^t_{\text{comb}} \) 
through a multi-layer perceptron (MLP) regressor to predict suitable gait parameters (\(\mathbf{g}\)), specifically, the robot's step height and hip splay. We train the regressor using the MSE loss, based on the training labels generated using the analysis of (\ref{eq:cost}) on the dataset.

The regressor contains two fully connected hidden layers with ReLU activation, while the output layer does not use any activation function. ReLU introduces nonlinearity, enabling the network to model complex relationships between the combined representation and gait parameters.

The model learns continuous mappings from sensory inputs to gait parameters, allowing it to generalize to unseen terrains by interpolating or extrapolating from training patterns. This approach ensures dynamic gait adaptation based on real-time sensory data, providing a robust solution for navigating complex terrains.
A dynamic window approach smoothens consecutive gait parameter predictions, ensuring stable and controlled gait transitions.

\section{Analysis and Results}

\subsection{Implementation}
We implemented \ours{} using PyTorch and trained all models on an NVIDIA A5000 GPU. For real-world experiments, we used the Ghost Vision 60 robot from Ghost Robotics, equipped with a front-facing wide-angle camera and an onboard Intel NUC 11 (Intel i7 CPU, NVIDIA RTX 2060 GPU). Our system runs in real-time at 60Hz. Path planning was implemented using \cite{sathyamoorthy2024mim}. We sample a 6-channel IMU and joint effort data, comprising 12 channels (one for each motor, measured in Nm), at 25 Hz. We combine these data streams to form the input time series data \(\mathbf{X}^t_{\text{TS}}\), resulting in \(n_{\text{ch}} = 18\). \(d_{\text{TS}} = 100\), is chosen empirically to capture the frequency components of the robot's cyclic gaits. \(d_m = 160\) based on \cite{franceschi2019unsupervised}.

We use a camera input, captured at 20 Hz, to generate the visual representation. Here, \(n_i = 224\), \(n_v = 49\), and \(d_e = 768\), as dictated by the ViT backbone \cite{he2022masked}. \(d_k = d_v = 160\), \(N_{sa} = 7\), and \(n_h = 4\), all chosen empirically to improve classification performance.

The MLP consists of two hidden layers: the first with 128 units and the second with 64 units. The final output layer has 2 units which correspond to the predicted gait parameters. All inputs to the overall system were normalized before being used both during training and testing.

\subsection{Dataset Collection}

\label{sec:Dataset}

We collected the dataset during a 9-hour period on a college campus, covering five distinct terrain types: hard surfaces (asphalt, concrete, brick), grass (lawns, soccer fields), dense vegetation, sand, and pebbles (including gravel). Step height values were varied between {0.03, 0.12, 0.21, 0.3} m, and hip splay values between {0.05, 0.09, 0.13, 0.17, 0.2} m to create 20 distinct gaits across these terrains. We recorded IMU data, joint effort data, and camera data, resulting in a dataset comprising approximately 120,000 images and 150,000 time-series samples.

We evaluated the cost function in (\ref{eq:cost}) for each pair of parameters in the dataset. The pair with the lowest cost in a given terrain is selected as the optimal gait parameter, which serves as the training label for that terrain. See Fig.\ref{fig:Params}.

\begin{figure}[!t]
      \centering
      
      \includegraphics[width=\columnwidth]{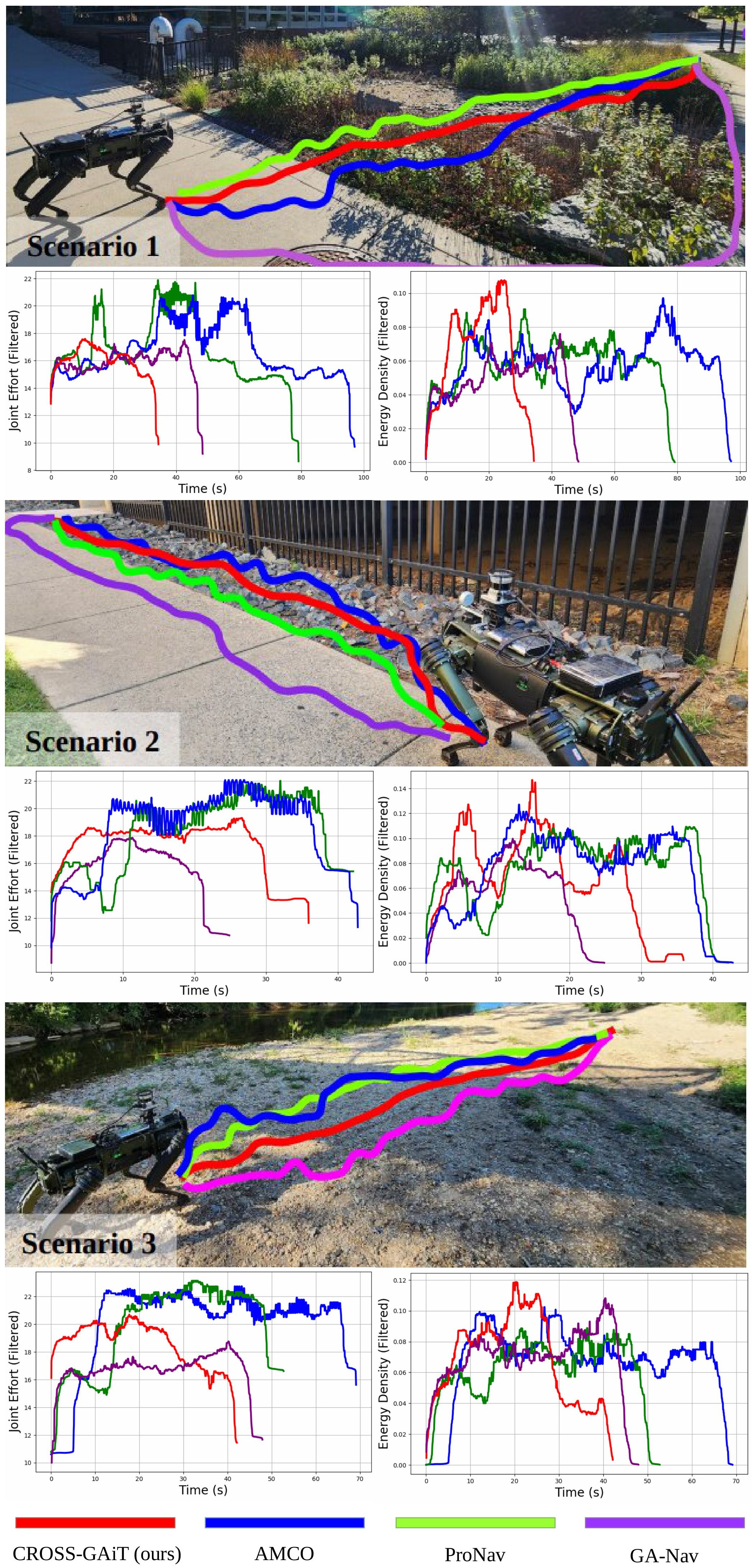}
      \caption {\small{ Robot trajectories in unstructured terrains using \ours{} and comparison methods along with corresponding Joint Effort (left) and IMU Energy Density (right) plots are shown below each scenario. \ours{} reaches the goal faster with lower joint effort, while maintaining comparable IMU energy density to other methods.
      }}
      \label{fig:traj_comparison}
      \vspace{-12pt}
\end{figure}

\begin{table}[h]
\renewcommand{\arraystretch}{1.25} 
\resizebox{\columnwidth}{!}{%
\begin{tabular}{ c c c c c } 
\hline
\textbf{Method} & \textbf{Accuracy (\%)} & \textbf{Precision (\%)} & \textbf{Recall (\%)} & \textbf{F1-Score (\%)} \\ [0.5ex] 
\hline

STERLING \cite{karnan2023sterling}  & 94.21 & 96.59 & 97.15 & 96.86 \\

\ours{} w/o Vision  & 93.48 & 93.49 & 93.47 & 93.41 \\

\ours{} w/o Prop.  & 94.19 & 94.10 & 95.23 & 94.16 \\

\ours{} w MLP  & 95.27 & 95.41 & 95.38 & 95.44 \\
\ours{} (ours) & \textbf{98.45} & \textbf{97.52} & \textbf{98.64} & \textbf{98.09} \\

\hline
\end{tabular}
}
\caption{\small{Comparison of our proposed method, \ours{} against STERLING \cite{karnan2023sterling}, along with ablation studies removing either vision or proprioceptive sensor modalities and an ablation replacing cross attention with MLP-based fusion, to evaluate the classification performance of learned representations using a Support Vector Classifier.}}
\label{tab:accuracy_comparison}
\vspace{-10pt}
\end{table}

\subsection{Comparison Methods and Metrics}

We compare our method against several existing approaches: 
STERLING \cite{karnan2023sterling}, which combines vision and proprioception for terrain navigation based on user preference (since the planner for STERLING is not publicly available, we evaluate the quality of fused embeddings through a terrain classification task using the dataset described in Section \ref{sec:Dataset}); 
AMCO \cite{elnoor2024amco}, which uses vision, proprioception, and battery data to compute traversability and switch between predefined gaits; ProNav \cite{elnoor2024pronav}, which relies on proprioception and battery data for traversability and gait switching; GA-Nav \cite{guan2022ga}, which leverages image segmentation for terrain understanding and traversability estimation; and the built-in planner from Ghost Robotics Vision60.

We also compare with ablation studies, testing variations with and without proprioception and vision-based inputs.

Our metrics for evaluation are:

\label{metric:GaitSelection}

\no \textbf{Success Rate} - The proportion of trials in which the robot successfully reaches its goal while maintaining stability (i.e., without collapsing) and avoiding collisions.

\no \textbf{Cumulative Joint Effort} - The total Joint Effort aggregated over time. A measure of the energy expenditure of the robot.

\no \textbf{RMS IMU Energy Density} - The root mean square of the aggregated squared accelerations and angular velocities measured by the IMU sensors across the x, y, and z axes, calculated over successful trials. Gravity is corrected for along the z-axis.

\no \textbf{Time to Reach Goal} - The time taken to reach the goal in seconds. Only successful attempts are counted.

\subsection{Testing Scenarios}

We evaluate our method's performance in real-world outdoor test scenarios featuring a combination of different terrains, including terrains out of the training dataset (e.g., mulch, irregular rocks, etc). Each scenario (see Fig. \ref{fig:cover-image} and Fig. \ref{fig:traj_comparison}) is tested with at least 10 trials. A trial is considered successful only if it navigates through the expected terrain. 

% \begin{itemize}
\no \textbf{Scenario 1} - Concrete, mulch, pebbles, and dense vegetation. (as in Fig. \ref{fig:cover-image}).

\no \textbf{Scenario 2} - Concrete and irregularly shaped rocky terrains.

\no \textbf{Scenario 3} - Sand, and pebbles with varying deformability.

\no \textbf{Scenario 4} - Concrete, mulch, gravel, tall and dense vegetation.
% \end{itemize}

\subsection{Analysis and Discussion}

We present our method results and comparisons qualitatively in Fig. \ref{fig:traj_comparison}, and quantitatively in Table \ref{tab:accuracy_comparison}, \ref{tab:comparison_table}. 

% \ours{} consistently outperforms all other methods in terms of Success Rate and Time to Reach Goal across all four scenarios. This is due to its ability to dynamically change the gait parameters, adapting smoothly and continuously across a spectrum of gait parameters rather than being restricted to a fixed set.

\ours{} consistently outperforms all other methods in Success Rate and Time to Reach Goal across all four scenarios. This superior performance stems from its ability to dynamically adjust gait parameters in real-time, enabling smooth and continuous adaptation across a spectrum of gaits rather than being constrained to a predefined set.

\begin{table}
\resizebox{\columnwidth}{!}{%
\begin{tabular}{ c c c c c c } 
\hline
\textbf{Metrics} & \textbf{Methods} & \multicolumn{1}{p{1cm}}{\centering \textbf{Scenario
1}} & \multicolumn{1}{p{1cm}}{\centering \textbf{Scenario
2}} & \multicolumn{1}{p{1cm}}{\centering \textbf{Scenario
3}} & \multicolumn{1}{p{1cm}}{\centering \textbf{Scenario
4}}\\ [0.5ex] 
\hline

\multirow{6}{*}{\rotatebox[origin=c]{0}{\makecell{\textbf{Success}\\\textbf{Rate (\%) $\uparrow$} }}} 
 & ProNAV \cite{elnoor2024pronav}  & 50 & 20 &  60 & 20   \\
 & AMCO \cite{elnoor2024amco} & 60 & 40 & \textbf{80} & 30 \\
 & GA-Nav \cite{guan2022ga}  & n/a & n/a & 60 & n/a \\
 &  \ours{} w/o Proprioception  & 80 & 60 & 70 & 50\\
 &  \ours{} w/o Vision  & 70 & 30 & 70 & 30 \\
 &  \ours{} (ours) & \textbf{90} &  \textbf{70} &  \textbf{80} &  \textbf{70}\\
\hline

\multirow{6}{*}{\rotatebox[origin=c]{0}{\makecell{\textbf{Cumulative} \\\textbf{Joint Effort $\downarrow$}\\\textbf{(Nm) \(\times 10^4\)} }}} 
 &  ProNAV \cite{elnoor2024pronav}  & 3.352 & 1.9842 & 2.6536 &  1.2772  \\
 & AMCO \cite{elnoor2024amco} & 4.1041 & 2.0263 & 3.5309 & 1.2145 \\
 & GA-Nav \cite{guan2022ga}  & 1.9807 & \textbf{0.9961} & 2.0945 & n/a \\
  &  \ours{} w/o Proprioception & 1.8942 & 1.9742 & 2.4576 & 1.2112\\
 &  \ours{} w/o Vision & 2.1213 & 2.3642 & 3.704 & 1.9823\\
 &  \ours{} (ours) & \textbf{1.4228} & 1.5848 $\dagger$ & \textbf{2.0084} & \textbf{1.1265}\\
\hline

\multirow{6}{*}{\rotatebox[origin=c]{0}{\makecell{\textbf{RMS IMU} \\\textbf{Energy Density $\downarrow$}}}} 
 &  ProNAV \cite{elnoor2024pronav}  & 0.092 & 0.137 & 0.099 &  0.108  \\
 & AMCO \cite{elnoor2024amco} & 0.091 $\dagger$ & 0.126 & 0.105 & 0.109 \\
 & GA-Nav \cite{guan2022ga}  & \textbf{0.081} & \textbf{0.093} & 0.0928 & n/a \\
  &  \ours{} w/o Proprioception & 0.093 & 0.106 & \textbf{0.912} & 0.102\\
  &  \ours{} w/o Vision & 0.095 & 0.112 & 0.0922 & 0.112\\
 &  \ours{} (ours) & 0.092 & 0.104 $\dagger$ & \textbf{0.0912} & \textbf{0.099}\\
\hline

\multirow{6}{*}{\rotatebox[origin=c]{0}{\makecell{\textbf{Time to }\\\textbf{Reach Goal $\downarrow$} \\\textbf{(s)}} }} 
 &  ProNAV \cite{elnoor2024pronav}  & 79.14 & 42.15 & 52.73 &  31.0  \\
 & AMCO \cite{elnoor2024amco} & 97.08 & 42.79 & 69.14 & 31.08 \\
 & GA-Nav \cite{guan2022ga}  & 48.5 & \textbf{24.85} & 47.92 & n/a \\
  &  \ours{} w/o Proprioception & 35.78 & 37.87 & 45.12 & 30.54\\
 &  \ours{} w/o Vision & 39.78 & 38.12 & 48.12 & 31.12\\
 &  \ours{} (ours) & \textbf{34.36} & 35.91 $\dagger$ & \textbf{42.13} & \textbf{29.00}\\
\hline
\end{tabular}
}
\caption{\small{Improved navigation performance of \ours{}, compared to other methods. We observe considerable improvement in all the four metrics. $\dagger$ denotes the best value when considering only successful trials.}
}
\label{tab:comparison_table}
\vspace{-18pt}
\end{table}

In \textbf{Scenario 1}, GA-Nav~\cite{guan2022ga} avoids the vegetated area by taking a longer route, while ProNav~\cite{elnoor2024pronav} and AMCO~\cite{elnoor2024amco} switch to predefined gaits with increased step height but still get stuck in the thick grass. In contrast, \ours{} adapts dynamically by further increasing step height to navigate the dense terrain efficiently, allowing it to reach the goal faster. \ours{} handled mulch regions as if it were a terrain between a hard surface and sand, demonstrating its ability to generalize to unseen terrains. Additionally, \ours{} reduces joint effort on harder surfaces by minimizing unnecessary leg lifting, improving energy efficiency.

In \textbf{Scenario 2}, both AMCO and ProNav switch to predefined gaits that increase both step height and hip splay while traversing the rocks. In contrast, \ours{} increases hip splay without significantly raising step height, resulting in lower RMS IMU energy density (less vibration) and lower joint effort. This leads to  and faster goal completion due to improved stability and fewer instabilities. GA-Nav avoids the rocky regions due to its fixed semantic cost map. This is not the expected behavior for a direct comparison in our context. 

In \textbf{Scenario 3}, AMCO and ProNav switch to predefined gaits with higher steps and a wider stance, leading to higher joint efforts. GA-Nav doesn’t adjust its gait, causing its legs to occasionally get stuck in the sand. In contrast, \ours{} adapts dynamically, selecting gait parameters that balance between sand and rocks, demonstrating strong generalization. As a result, \ours{} outperforms all other methods in terms of all defined metrics. GA-Nav gets stuck before being able to navigate through the grass. 

In \textbf{Scenario 4}, ProNav and AMCO switch to predefined gaits with higher steps, but they still get stuck in the bushes, leading to low success rates. In contrast, \ours{} avoids getting stuck by dynamically lifting its legs high enough to clear obstacles like bushes, resulting in lower cumulative joint effort and a significantly higher success rate than the other methods.

\no\textbf{Comparison on Embedding space:} In Table \ref{tab:accuracy_comparison}, we compare the terrain classification performance of the embeddings generated by CROSS-GAiT and STERLING~\cite{karnan2023sterling}. CROSS-GAiT achieves higher accuracy, precision, recall, and F1-score. By combining visual and time-series data through cross-attention, CROSS-GAiT generates richer latent representations, enhancing the classifier's ability to distinguish between diverse terrains.

\rev {\no\textbf{Ablation on Cross-attention:}  We replace our cross-attention-based fusion with a concatenation followed by an MLP layer, following the approach in \cite{karnan2023sterling}, as shown in Table \ref{tab:accuracy_comparison}, to assess the benefits of cross-attention. Our results show that cross-attention improves terrain classification accuracy,  precision, recall, and F1-score compared to concatenation-based fusion. This improvement stems from cross-attention’s ability to dynamically align and weight features across modalities, enabling richer and more context-aware representations, whereas MLP-based concatenation applies static transformations, limiting its adaptability to multimodal variations.}

\rev{ \no \textbf{Real-time Inference: } Our lightweight model achieves an inference rate around 60 Hz on a mobile computing platform equipped with an Intel i7 CPU and NVIDIA RTX 2060 GPU, demonstrating its efficiency for real-time continuous gait adaptation. This validates its feasibility for real-world deployment on mobile robots, ensuring responsiveness and adaptability in dynamic environments.}

\section{Conclusions, Limitations, and Future Work}
In this paper, we present a novel framework to enhance robot locomotion across diverse terrains using multimodal sensory inputs: IMU readings, joint forces, and visual inputs. Our approach leverages cross-attention to fuse these inputs, generating a comprehensive terrain representation that allows dynamic adjustment of gait parameters like step height and hip splay. This enables the robot to navigate complex environments with improved stability, energy efficiency, and reduced time to reach the goal.

The key contributions of our method include the fusion of multimodal inputs through cross-attention, creating a richer terrain understanding for more informed gait adjustments. Additionally, we employ a regressor that allows continuous and adaptive adjustment of gait parameters, enabling generalization to unseen terrains, unlike predefined gait strategies. While the method performs well, optimizing the latent representations for all terrain types remains a challenge. Future work will focus on refining these representations, incorporating additional sensory data, and using offline reinforcement learning techniques to learn optimal gaits without the need for labeled data, further enhancing the system's ability to adapt to diverse terrains.

\bibliographystyle{IEEEtran}
\bibliography{References}

\begin{thebibliography}{10}
\providecommand{\url}[1]{#1}
\csname url@rmstyle\endcsname
\providecommand{\newblock}{\relax}
\providecommand{\bibinfo}[2]{#2}
\providecommand\BIBentrySTDinterwordspacing{\spaceskip=0pt\relax}
\providecommand\BIBentryALTinterwordstretchfactor{4}
\providecommand\BIBentryALTinterwordspacing{\spaceskip=\fontdimen2\font plus
\BIBentryALTinterwordstretchfactor\fontdimen3\font minus \fontdimen4\font\relax}
\providecommand\BIBforeignlanguage[2]{{%
\expandafter\ifx\csname l@#1\endcsname\relax
\typeout{** WARNING: IEEEtran.bst: No hyphenation pattern has been}%
\typeout{** loaded for the language `#1'. Using the pattern for}%
\typeout{** the default language instead.}%
\else
\language=\csname l@#1\endcsname
\fi
#2}}

\bibitem{miki2022learning}
T.~Miki, J.~Lee, J.~Hwangbo, L.~Wellhausen, V.~Koltun, and M.~Hutter, ``Learning robust perceptive locomotion for quadrupedal robots in the wild,'' \emph{Science robotics}, vol.~7, no.~62, p. eabk2822, 2022.

\bibitem{frey2022locomotion}
J.~Frey, D.~Hoeller, S.~Khattak, and M.~Hutter, ``Locomotion policy guided traversability learning using volumetric representations of complex environments,'' in \emph{2022 IEEE/RSJ International Conference on Intelligent Robots and Systems (IROS)}.\hskip 1em plus 0.5em minus 0.4em\relax IEEE, 2022, pp. 5722--5729.

\bibitem{frey2023fast}
J.~Frey, M.~Mattamala, N.~Chebrolu, C.~Cadena, M.~Fallon, and M.~Hutter, ``Fast traversability estimation for wild visual navigation,'' \emph{arXiv preprint arXiv:2305.08510}, 2023.

\bibitem{sathyamoorthy2024mim}
A.~J. Sathyamoorthy, K.~Weerakoon, M.~Elnoor, M.~Russell, J.~Pusey, and D.~Manocha, ``Mim: Indoor and outdoor navigation in complex environments using multi-layer intensity maps,'' in \emph{2024 IEEE International Conference on Robotics and Automation (ICRA)}.\hskip 1em plus 0.5em minus 0.4em\relax IEEE, 2024, pp. 10\,917--10\,924.

\bibitem{agarwal2023legged}
A.~Agarwal, A.~Kumar, J.~Malik, and D.~Pathak, ``Legged locomotion in challenging terrains using egocentric vision,'' in \emph{Conference on robot learning}.\hskip 1em plus 0.5em minus 0.4em\relax PMLR, 2023, pp. 403--415.

\bibitem{karnan2023sterling}
H.~Karnan, E.~Yang, D.~Farkash, G.~Warnell, J.~Biswas, and P.~Stone, ``Sterling: Self-supervised terrain representation learning from unconstrained robot experience,'' in \emph{7th Annual Conference on Robot Learning}, 2023.

\bibitem{fu2021minimizing}
Z.~Fu, A.~Kumar, J.~Malik, and D.~Pathak, ``Minimizing energy consumption leads to the emergence of gaits in legged robots,'' \emph{arXiv preprint arXiv:2111.01674}, 2021.

\bibitem{roy2012effects}
S.~S. Roy and D.~K. Pratihar, ``Effects of turning gait parameters on energy consumption and stability of a six-legged walking robot,'' \emph{Robotics and Autonomous Systems}, vol.~60, no.~1, pp. 72--82, 2012.

\bibitem{hutter2016anymal}
M.~Hutter, C.~Gehring, D.~Jud, A.~Lauber, C.~D. Bellicoso, V.~Tsounis, J.~Hwangbo, K.~Bodie, P.~Fankhauser, M.~Bloesch, \emph{et~al.}, ``Anymal-a highly mobile and dynamic quadrupedal robot,'' in \emph{2016 IEEE/RSJ international conference on intelligent robots and systems (IROS)}.\hskip 1em plus 0.5em minus 0.4em\relax IEEE, 2016, pp. 38--44.

\bibitem{elnoor2024amco}
M.~Elnoor, K.~Weerakoon, A.~J. Sathyamoorthy, T.~Guan, V.~Rajagopal, and D.~Manocha, ``Amco: Adaptive multimodal coupling of vision and proprioception for quadruped robot navigation in outdoor environments,'' \emph{arXiv preprint arXiv:2403.13235}, 2024.

\bibitem{elnoor2024pronav}
M.~Elnoor, A.~J. Sathyamoorthy, K.~Weerakoon, and D.~Manocha, ``Pronav: Proprioceptive traversability estimation for legged robot navigation in outdoor environments,'' \emph{IEEE Robotics and Automation Letters}, 2024.

\bibitem{guan2022ga}
T.~Guan, D.~Kothandaraman, R.~Chandra, A.~J. Sathyamoorthy, K.~Weerakoon, and D.~Manocha, ``Ga-nav: Efficient terrain segmentation for robot navigation in unstructured outdoor environments,'' \emph{IEEE Robotics and Automation Letters}, vol.~7, no.~3, pp. 8138--8145, 2022.

\bibitem{yu2021visual}
W.~Yu, D.~Jain, A.~Escontrela, A.~Iscen, P.~Xu, E.~Coumans, S.~Ha, J.~Tan, and T.~Zhang, ``Visual-locomotion: Learning to walk on complex terrains with vision,'' in \emph{5th Annual Conference on Robot Learning}, 2021.

\bibitem{li2023seeing}
A.~Li, C.~Yang, J.~Frey, J.~Lee, C.~Cadena, and M.~Hutter, ``Seeing through the grass: Semantic pointcloud filter for support surface learning,'' \emph{IEEE Robotics and Automation Letters}, 2023.

\bibitem{xue2018deep}
J.~Xue, H.~Zhang, and K.~Dana, ``Deep texture manifold for ground terrain recognition,'' in \emph{Proceedings of the IEEE Conference on Computer Vision and Pattern Recognition}, 2018, pp. 558--567.

\bibitem{sathyamoorthy2023vern}
A.~J. Sathyamoorthy, K.~Weerakoon, T.~Guan, M.~Russell, D.~Conover, J.~Pusey, and D.~Manocha, ``Vern: Vegetation-aware robot navigation in dense unstructured outdoor environments,'' in \emph{2023 IEEE/RSJ International Conference on Intelligent Robots and Systems (IROS)}.\hskip 1em plus 0.5em minus 0.4em\relax IEEE, 2023, pp. 11\,233--11\,240.

\bibitem{weerakoon2022graspe}
K.~Weerakoon, A.~J. Sathyamoorthy, J.~Liang, T.~Guan, U.~Patel, and D.~Manocha, ``Graspe: Graph based multimodal fusion for robot navigation in unstructured outdoor environments,'' \emph{arXiv preprint arXiv:2209.05722}, 2022.

\bibitem{dey2022prepare}
S.~Dey, D.~Fan, R.~Schmid, A.~Dixit, K.~Otsu, T.~Touma, A.~F. Schilling, and A.-A. Agha-Mohammadi, ``Prepare: Predictive proprioception for agile failure event detection in robotic exploration of extreme terrains,'' in \emph{2022 IEEE/RSJ International Conference on Intelligent Robots and Systems (IROS)}.\hskip 1em plus 0.5em minus 0.4em\relax IEEE, 2022, pp. 4338--4343.

\bibitem{kumar2021rma}
A.~Kumar, Z.~Fu, D.~Pathak, and J.~Malik, ``Rma: Rapid motor adaptation for legged robots,'' \emph{arXiv preprint arXiv:2107.04034}, 2021.

\bibitem{sikand2022visual}
K.~S. Sikand, S.~Rabiee, A.~Uccello, X.~Xiao, G.~Warnell, and J.~Biswas, ``Visual representation learning for preference-aware path planning,'' in \emph{2022 International Conference on Robotics and Automation (ICRA)}.\hskip 1em plus 0.5em minus 0.4em\relax IEEE, 2022, pp. 11\,303--11\,309.

\bibitem{castro2023does}
M.~G. Castro, S.~Triest, W.~Wang, J.~M. Gregory, F.~Sanchez, J.~G. Rogers, and S.~Scherer, ``How does it feel? self-supervised costmap learning for off-road vehicle traversability,'' in \emph{2023 IEEE International Conference on Robotics and Automation (ICRA)}.\hskip 1em plus 0.5em minus 0.4em\relax IEEE, 2023, pp. 931--938.

\bibitem{wellhausen2019should}
L.~Wellhausen, A.~Dosovitskiy, R.~Ranftl, K.~Walas, C.~Cadena, and M.~Hutter, ``Where should i walk? predicting terrain properties from images via self-supervised learning,'' \emph{IEEE Robotics and Automation Letters}, vol.~4, no.~2, pp. 1509--1516, 2019.

\bibitem{kahn2021badgr}
G.~Kahn, P.~Abbeel, and S.~Levine, ``Badgr: An autonomous self-supervised learning-based navigation system,'' \emph{IEEE Robotics and Automation Letters}, vol.~6, no.~2, pp. 1312--1319, 2021.

\bibitem{yao2022rca}
X.~Yao, J.~Zhang, and J.~Oh, ``Rca: Ride comfort-aware visual navigation via self-supervised learning,'' in \emph{2022 IEEE/RSJ International Conference on Intelligent Robots and Systems (IROS)}.\hskip 1em plus 0.5em minus 0.4em\relax IEEE, 2022, pp. 7847--7852.

\bibitem{liu2024contrastive}
X.~Liu, H.~Chen, and H.~Chen, ``Contrastive learning-based attribute extraction method for enhanced terrain classification,'' in \emph{2024 IEEE International Conference on Robotics and Automation (ICRA)}.\hskip 1em plus 0.5em minus 0.4em\relax IEEE, 2024, pp. 5644--5650.

\bibitem{matthis2018gaze}
J.~S. Matthis, J.~L. Yates, and M.~M. Hayhoe, ``Gaze and the control of foot placement when walking in natural terrain,'' \emph{Current Biology}, vol.~28, no.~8, pp. 1224--1233, 2018.

\bibitem{lewis-gait}
M.~Lewis, C.~Waltz, L.~Scelina, K.~Scelina, K.~Owen, K.~Hastilow, E.~Zimmerman, A.~Rosenfeldt, M.~Koop, and J.~Alberts, ``Gait patterns during overground and virtual omnidirectional treadmill walking,'' \emph{Journal of NeuroEngineering and Rehabilitation}, vol.~21, 02 2024.

\bibitem{schilling2017geometric}
F.~Schilling, X.~Chen, J.~Folkesson, and P.~Jensfelt, ``Geometric and visual terrain classification for autonomous mobile navigation,'' in \emph{2017 IEEE/RSJ International Conference on Intelligent Robots and Systems (IROS)}.\hskip 1em plus 0.5em minus 0.4em\relax IEEE, 2017, pp. 2678--2684.

\bibitem{fahmi2022vital}
S.~Fahmi, V.~Barasuol, D.~Esteban, O.~Villarreal, and C.~Semini, ``Vital: Vision-based terrain-aware locomotion for legged robots,'' \emph{IEEE Transactions on Robotics}, vol.~39, no.~2, pp. 885--904, 2022.

\bibitem{aladem2019evaluation}
M.~Aladem, S.~Baek, and S.~A. Rawashdeh, ``Evaluation of image enhancement techniques for vision-based navigation under low illumination,'' \emph{Journal of Robotics}, vol. 2019, no.~1, p. 5015741, 2019.

\bibitem{sathyamoorthy2022terrapn}
A.~J. Sathyamoorthy, K.~Weerakoon, T.~Guan, J.~Liang, and D.~Manocha, ``Terrapn: Unstructured terrain navigation using online self-supervised learning,'' in \emph{2022 IEEE/RSJ International Conference on Intelligent Robots and Systems (IROS)}.\hskip 1em plus 0.5em minus 0.4em\relax IEEE, 2022, pp. 7197--7204.

\bibitem{bardes2021vicreg}
A.~Bardes, J.~Ponce, and Y.~LeCun, ``Vicreg: Variance-invariance-covariance regularization for self-supervised learning,'' \emph{arXiv preprint arXiv:2105.04906}, 2021.

\bibitem{zhu2024cross}
S.~Zhu, D.~Li, Y.~Liu, N.~Xu, and H.~Zhao, ``Cross anything: General quadruped robot navigation through complex terrains,'' \emph{arXiv preprint arXiv:2407.16412}, 2024.

\bibitem{loquercio2023learning}
A.~Loquercio, A.~Kumar, and J.~Malik, ``Learning visual locomotion with cross-modal supervision,'' in \emph{IEEE International Conference on Robotics and Automation (ICRA)}.\hskip 1em plus 0.5em minus 0.4em\relax IEEE, 2023, pp. 7295--7302.

\bibitem{fu2022coupling}
Z.~Fu, A.~Kumar, A.~Agarwal, H.~Qi, J.~Malik, and D.~Pathak, ``Coupling vision and proprioception for navigation of legged robots,'' in \emph{Proceedings of the IEEE/CVF Conference on Computer Vision and Pattern Recognition}, 2022, pp. 17\,273--17\,283.

\bibitem{bahaduri2023multimodal}
B.~Bahaduri, Z.~Ming, F.~Feng, and A.~Mokraou, ``Multimodal transformer using cross-channel attention for object detection in remote sensing images,'' \emph{arXiv preprint arXiv:2310.13876}, 2023.

\bibitem{wei2020multi}
X.~Wei, T.~Zhang, Y.~Li, Y.~Zhang, and F.~Wu, ``Multi-modality cross attention network for image and sentence matching,'' in \emph{Proceedings of the IEEE/CVF conference on computer vision and pattern recognition}, 2020, pp. 10\,941--10\,950.

\bibitem{li2024crossfuse}
H.~Li and X.-J. Wu, ``Crossfuse: A novel cross attention mechanism based infrared and visible image fusion approach,'' \emph{Information Fusion}, vol. 103, p. 102147, 2024.

\bibitem{praveen2024recursive}
R.~G. Praveen and J.~Alam, ``Recursive cross-modal attention for multimodal fusion in dimensional emotion recognition,'' \emph{arXiv preprint arXiv:2403.13659}, 2024.

\bibitem{limultimodalcatt}
K.~Li, L.~Xu, C.~Zhu, and K.~Zhang, ``A multimodal graph recommendation method based on cross-attention fusion,'' \emph{Mathematics}, vol.~12, p. 2353, 07 2024.

\bibitem{zhao2024bronchocopilot}
J.~Zhao, H.~Chen, Q.~Tian, J.~Chen, B.~Yang, and H.~Liu, ``Bronchocopilot: Towards autonomous robotic bronchoscopy via multimodal reinforcement learning,'' \emph{arXiv preprint arXiv:2403.01483}, 2024.

\bibitem{bengio2013representation}
Y.~Bengio, A.~Courville, and P.~Vincent, ``Representation learning: A review and new perspectives,'' \emph{IEEE transactions on pattern analysis and machine intelligence}, vol.~35, no.~8, pp. 1798--1828, 2013.

\bibitem{he2022masked}
K.~He, X.~Chen, S.~Xie, Y.~Li, P.~Doll{\'a}r, and R.~Girshick, ``Masked autoencoders are scalable vision learners,'' in \emph{Proceedings of the IEEE/CVF conference on computer vision and pattern recognition}, 2022, pp. 16\,000--16\,009.

\bibitem{dosovitskiy2020image}
A.~Dosovitskiy, ``An image is worth 16x16 words: Transformers for image recognition at scale,'' \emph{arXiv preprint arXiv:2010.11929}, 2020.

\bibitem{ma2021ecg}
H.~Ma, C.~Chen, Q.~Zhu, H.~Yuan, L.~Chen, and M.~Shu, ``An ecg signal classification method based on dilated causal convolution,'' \emph{Computational and Mathematical Methods in Medicine}, vol. 2021, no.~1, p. 6627939, 2021.

\bibitem{franceschi2019unsupervised}
J.-Y. Franceschi, A.~Dieuleveut, and M.~Jaggi, ``Unsupervised scalable representation learning for multivariate time series,'' \emph{Advances in neural information processing systems}, vol.~32, 2019.

\bibitem{rajan2022cross}
V.~Rajan, A.~Brutti, and A.~Cavallaro, ``Is cross-attention preferable to self-attention for multi-modal emotion recognition?'' in \emph{ICASSP 2022-2022 IEEE International Conference on Acoustics, Speech and Signal Processing (ICASSP)}.\hskip 1em plus 0.5em minus 0.4em\relax IEEE, 2022, pp. 4693--4697.

\bibitem{vaswani2017attention}
A.~Vaswani, ``Attention is all you need,'' \emph{Advances in Neural Information Processing Systems}, 2017.

\bibitem{schroff2015facenet}
F.~Schroff, D.~Kalenichenko, and J.~Philbin, ``Facenet: A unified embedding for face recognition and clustering,'' in \emph{Proceedings of the IEEE conference on computer vision and pattern recognition}, 2015, pp. 815--823.

\bibitem{khosla2020supervised}
P.~Khosla, P.~Teterwak, C.~Wang, A.~Sarna, Y.~Tian, P.~Isola, A.~Maschinot, C.~Liu, and D.~Krishnan, ``Supervised contrastive learning,'' \emph{Advances in neural information processing systems}, vol.~33, pp. 18\,661--18\,673, 2020.

\end{thebibliography}
\clearpage

% Switch to single-column mode
\onecolumn

% Start Appendix section
\appendix

\section{Appendix}
\subsection{Hyperparameters Used During Training}

The training process for the CROSS-GAiT model involves multiple stages, where different components are trained separately and then combined. The common hyperparameters for each component in CROSS-GAiT are summarized table \ref{tab:common-hyps}. Furthermore, the specific hyperparameters for each method are summarized in table \ref{tab:individual_hyperparameters}.

\renewcommand{\arraystretch}{1.5} % Adjust row height (1.5x the default)
\begin{table}[h!]
    \centering
    \begin{tabularx}{\textwidth}{|X|X|X|X|X|}
    \hline
    \textbf{Hyperparameter} & \textbf{Masked Autoencoder} & \textbf{Dilated Causal Conv. Encoder} & \textbf{Cross-Attention Fusion} & \textbf{Fully Connected Network} \\
    \hline
    \textbf{Learning Rate} & 0.00025 & 0.0001 & 0.00005 & 0.0001 \\
    \hline
    \textbf{Batch Size} & 128 & 256 & 1024 & 1024 \\
    \hline
    \textbf{Optimizer} & AdamW & Adam & AdamW & AdamW \\
    \hline
    \textbf{Epochs} & 400 & 10 & 100 & 5 \\
    % \hline
    % \textbf{Learning Rate Scheduler} & \raggedright Cosine Annealing with Restart and Warmup& \raggedright Cosine Annealing and warmup & 100 & 50 \\ % Left-aligned
    \hline
    \end{tabularx}
    \caption{Common hyperparameters across different components of the CROSS-GAiT model.}
    \label{tab:common-hyps}
\end{table}
\vspace{-10pt}

\begin{table}[h!]
    \centering
    \begin{minipage}[t]{0.22\textwidth} % Slightly reduced width
        \centering
        \textbf{Masked Autoencoder}\newline
        \vspace{0pt} % Adjust this value for desired spacing
        \begin{tabular}{|p{2cm}|>{\centering\arraybackslash}p{1cm}|} % Centering values
        \hline
        \textbf{Hyperparameter} & \textbf{Value} \\
        \hline
        Embedding Dimension & 768 \\
        \hline
        Masking Ratio & 0.75 \\
        \hline
        Weight Decay & 0.05 \\
        \hline
        \end{tabular}
    \end{minipage}%
    \hspace{0.03\textwidth} % Adjust spacing between minipages
    \begin{minipage}[t]{0.22\textwidth} % Slightly reduced width
        \centering
        \textbf{Dilated Causal Conv. Encoder}
        \newline
        \vspace{1pt} % Adjust this value for desired spacing
        \begin{tabular}{|p{2cm}|>{\centering\arraybackslash}p{1cm}|} % Centering values
        \hline
        \textbf{Hyperparameter} & \textbf{Value} \\
        \hline
        Dilation Rate & 2 \\
        \hline
        Kernel Size & 3 \\
        \hline
        \end{tabular}
    \end{minipage}%
    \hspace{0.03\textwidth} % Adjust spacing between minipages
    \begin{minipage}[t]{0.22\textwidth} % Slightly reduced width
        \centering
        \textbf{Cross-Attention}
        \newline
        \vspace{2pt} % Adjust this value for desired spacing
        \begin{tabular}{|p{2cm}|>{\centering\arraybackslash}p{1cm}|} % Centering values
        \hline
        \textbf{Hyperparameter} & \textbf{Value} \\
        \hline
        Embedding Dimension & 160 \\
        \hline
        Dropout & 0.1 \\
        \hline
        Weight Decay & 0.01 \\
        \hline
        \raggedright Sup Con tempereature & 0.05 \\
        \hline
        \end{tabular}
    \end{minipage}%
    \hspace{0.03\textwidth} % Adjust spacing between minipages
    \caption{Individual hyperparameters for each component of the CROSS-GAiT model.}
    \label{tab:individual_hyperparameters}
\end{table}

\subsection{Dynamic Window for Gait Adaptation}

To ensure smooth transitions in gait parameters (step height and hip splay) while maintaining control over the robot's movements, a dynamic window mechanism is implemented. This mechanism limits the changes in step height and hip splay per iteration, as well as enforcing maximum and minimum values for both parameters to avoid motor control issues caused by hardware limitations.

\subsubsection{Maximum and Minimum Values}
The system enforces both maximum and minimum allowable values for step height and hip splay to avoid exceeding motor capabilities, ensuring safe and controllable movement.

\begin{itemize}
    \item \textbf{Maximum Step Height:} 0.3 meters
    \item \textbf{Minimum Step Height:} 0.03 meters
    \item \textbf{Maximum Hip Splay:} 0.2 meters
    \item \textbf{Minimum Hip Splay:} 0.05 meters
\end{itemize}

\subsubsection{Maximum Change Per Iteration}
The dynamic window also restricts the change in both step height and hip splay per iteration, allowing the system to adapt gradually without causing instability due to sudden shifts in gait.

\begin{itemize}
    \item \textbf{Maximum Step Height Change Per Iteration:} 0.01 meters
    \item \textbf{Maximum Hip Splay Change Per Iteration:} 0.01 meters
\end{itemize}

\subsection{Data Collection Process (extending Section \ref{sec:Dataset})}

The data was collected under various lighting conditions, such as bright sunlight, overcast skies, and shaded areas. These varying conditions were crucial for ensuring that the visual model could generalize across different environments and remain robust when encountering challenges like shadows or direct sunlight, which could otherwise affect the perception of terrain.

Since the data collection was restricted to well-defined patches of specific terrain types (e.g., grass, hard surfaces, sand, rocks, and pebbles), there was no need to manually label individual images. The entire segment of images collected in a particular terrain zone was uniformly labeled, simplifying the process while maintaining data quality.

Additionally, to capture the full range of the robot's dynamics, the data was recorded at various walking speeds. This approach allowed the model to learn how different gaits performed across a range of speeds, from slow and deliberate movements to faster, more dynamic gaits. These variations also enriched the IMU and joint effort data, providing a more comprehensive representation of the robot’s interaction with diverse terrains.

\end{document}